\documentclass{article}

\usepackage[main, final]{neurips_2026}

\usepackage[utf8]{inputenc}%
\usepackage[T1]{fontenc}%

\usepackage{url}%
\usepackage{booktabs}%
\usepackage{amsfonts}%
\usepackage{amssymb}
\usepackage{graphicx}%
\usepackage{nicefrac}%
\usepackage{microtype}%
\usepackage{xcolor}%
\usepackage{amsmath}
\usepackage{multirow}
\usepackage{multicol}
\usepackage{tcolorbox}
\usepackage{subfigure}
\usepackage{makecell}
\usepackage{placeins}
\usepackage{color, colortbl}
\definecolor{citecolor}{HTML}{bda6ce}
\usepackage[breaklinks=true,bookmarks=false, colorlinks=true, linkcolor=ForestGreen,bookmarks=false, citecolor=citecolor]{hyperref}
\usepackage{natbib}
\setcitestyle{numbers,square,comma}
\usepackage{wrapfig}

\usepackage{tcolorbox}
\usepackage{placeins}
\tcbset{
  myboxstyle/.style={
    colback=gray!5,%
    colframe=gray!80,%
    boxrule=0.5pt,%
    arc=4pt,%
    outer arc=4pt,
    boxsep=5pt,%
    left=5pt,%
    right=5pt,%
    top=5pt,%
    bottom=5pt,%
  }
}
\usepackage{caption}

\definecolor{light-gray}{gray}{0.6}
\definecolor{front-color}{HTML}{F5FFFA}
\definecolor{Gray}{gray}{0.93}

\newcommand{\best}[1]{\textbf{\underline{#1}}} 
\newcommand{\sbest}[1]{\textbf{#1}}    
 
\definecolor{ForestGreen}{RGB}{116, 69, 119}
\definecolor{LighterGray}{gray}{0.93}
\definecolor{LightPurple}{RGB}{180, 211, 217}

\newcommand{\thickhline}{\Xhline{3\arrayrulewidth}}

\newcommand{\custompar}[1]{
  \par
  \vspace{2pt}
  \noindent\textbf{#1}
}

\newcommand{\startexcludefrom}[1]{
    \addtocontents{toc}{\protect\setcounter{tocdepth}{-1}}%
    #1
    \addtocontents{toc}{\protect\setcounter{tocdepth}{3}}%
}

\title{\raisebox{-0.2em}{\includegraphics[height=1.15em]{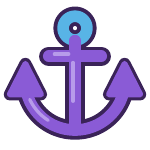}}~Mental Imagery That Matters: \\ Curing Latent Collapse in Multimodal Reasoning}

\author{
Mi Luo\textsuperscript{1}\space\space\space 
Rohan Dugad\textsuperscript{1}\space\space\space 
Zihui Xue\textsuperscript{1}\space\space\space 
Alex Dimakis\textsuperscript{2, 3}\space\space\space 
Kristen Grauman\textsuperscript{1}\space\space\space
\vspace{1mm} \\
\textsuperscript{1}The University of Texas at Austin\qquad 
\textsuperscript{2}UC Berkeley\qquad 
\textsuperscript{3}Bespoke Labs \vspace{1mm}\\
}

\begin{document}

\maketitle

\startexcludefrom{\begin{abstract}
Mental simulation, the ability to reason by constructing and updating internal visual states, is central to flexible human cognition, but remains elusive in current multimodal large language models. Latent multimodal reasoning offers a promising route toward this capability by interleaving text with continuous visual latent blocks, allowing models to maintain internal visual workspaces without decoding them into pixels. In this paper, we ask whether these latent visual states actually matter. Through a direct perturbation analysis across representative latent-reasoning MLLMs, we find that they often do not: dropping, replacing, or shuffling generated visual latents has only minor effects on the model's own final-answer probability, while equivalent perturbations to the linguistic reasoning trace are substantially more damaging. We call this failure mode \emph{Mental Simulation Collapse} (MSC): the model appears to perform interleaved visual reasoning, but routes most answer-relevant computation through text. We further diagnose that MSC follows naturally from existing training objectives: visual alignment can make generated latents resemble helper-image embeddings, but it does not require the decoder to rely on them; subsequent text-only supervision further leaves the semantic role of the latent pathway unidentifiable. To address this, we introduce \textbf{AnchorSim} (\textbf{Anchor}ed Mental \textbf{Sim}ulation), a contrastive distillation framework that trains models to make their generated visual latents consequential. Across multiple visual reasoning benchmarks, AnchorSim improves performance, while diagnostic re-evaluation shows that visual perturbations now meaningfully affect predictions. Our results expose a fundamental failure mode of latent multimodal reasoning and demonstrate that mental simulation must be anchored to model behavior in order to matter.{\renewcommand{\thefootnote}{\textcolor{magenta}{\arabic{footnote}}}\footnote{Project page: \url{https://vision.cs.utexas.edu/projects/AnchorSim/}}}
\end{abstract}
\section{Introduction}
\label{sec:intro}
Human visual reasoning often unfolds as an internal simulation rather than a purely linguistic process~\cite{nanay1997mental}. When people solve tasks involving spatial layouts, object transformations, physical dynamics, or navigation, they construct and update visual states in their mind: rotating a shape, tracing a path through a maze, or imagining how one object affects another~\cite{vjepa, vjepa2}. These intermediate states are not merely memories retrieved once and verbalized. They are manipulated, inspected, and revised over time~\cite{richardson2013mental}. For multimodal large language models, this suggests a natural goal: visual reasoning should not have to be compressed entirely into text~\cite{yang2024video}. Models should maintain and consult task-relevant visual states throughout their reasoning trajectory.

\begin{figure*}[!t]
    \centering
    \includegraphics[width=\linewidth]{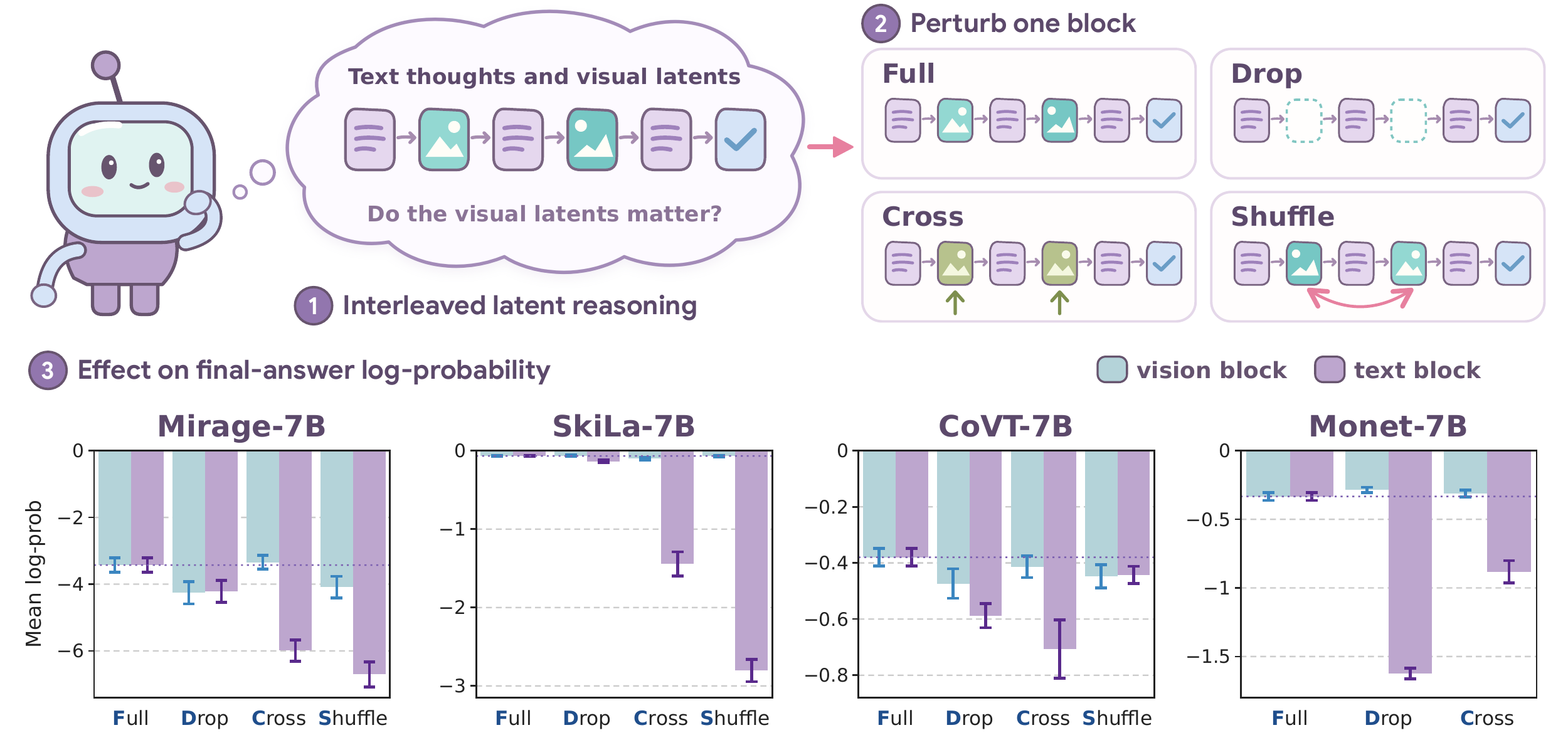}
    \caption{
      \textbf{Vision vs.\ text perturbation impact across four multimodal latent-reasoning models.} Each panel reports the mean log-probability (±1 SEM) of the model's final-answer span. We compare four conditions applied independently to the \emph{vision block} (blue) and the \emph{reasoning text block} (purple): \textbf{F}ull (unchanged), \textbf{D}rop (removed), \textbf{C}ross (replaced from another sample), and \textbf{S}huffle (permuted). Step 2 shows them on the vision block; the charts apply them to both blocks in turn. A larger drop from \textbf{F}ull indicates that the perturbed block carries more answer-relevant information. In most conditions, perturbing the text block degrades the answer substantially more than perturbing the visual block, indicating that computation is routed primarily through the linguistic stream.
    }
    \label{fig:vision_vs_text_all_models}
\end{figure*}

Recent work has moved toward this goal through \emph{interleaved multimodal reasoning}, where a model alternates between natural-language reasoning and generated visual representations. Some approaches make these representations explicit by producing intermediate images or video frames~\cite{wu2024vila, liu2025javisgpt, deng2025bagel, zebra, gu2025thinkmorph, mvot}. While interpretable, such methods are computationally expensive and require substantial image-generation machinery. Others rely on external visual tools, such as drawing, cropping, detection, or code execution, but are limited by the operations exposed through the tool interface~\cite{zhang2025thinking, yang2025longvt, fan2025grit, r3, ou2025bridging, zheng2025deepeyes, hong2025deepeyesv2}. A more lightweight alternative is to reason directly in latent space~\cite{mirage, lvr, covt, skila, monet, ilvr}. Instead of rendering pixels, the model emits continuous visual latent blocks between text spans, producing trajectories of the form $\langle \textcolor[HTML]{8A63B0}{\texttt{text}}_0, \textcolor[HTML]{2F9C95}{\texttt{vis}}_0, \textcolor[HTML]{8A63B0}{\texttt{text}}_1, \textcolor[HTML]{2F9C95}{\texttt{vis}}_1,\ldots,\textcolor[HTML]{4E86B8}{\texttt{answer}}\rangle$. In principle, these latent blocks could serve as compact mental images: internal visual workspaces that preserve information too cumbersome to express in language, and that later reasoning can use. This premise assumes that generated visual latents actually matter. If answers rely entirely on surrounding text, the model is merely decorating a linguistic chain-of-thought with inert visual placeholders. We therefore ask: do latent-reasoning MLLMs causally use generated visual states to produce the final answer?

Surprisingly, for existing systems, the answer is largely no. Evaluating four recent latent-reasoning MLLMs by perturbing their generated trajectories reveals that dropping, replacing, or shuffling visual latents barely affects the final-answer probability. Conversely, equivalent perturbations to the linguistic trace are highly damaging, as shown in Fig.~\ref{fig:vision_vs_text_all_models}. These visual states act as decorative placeholders rather than load-bearing computations, a failure mode we term \emph{Mental Simulation Collapse} (MSC). We further diagnose that MSC follows naturally from existing training pipelines~\cite{mirage, ilvr}. While initial alignment makes latents resemble helper images, subsequent text-only supervision only evaluates the final text distribution. Consequently, a decoder can achieve high likelihoods by routing information entirely through the linguistic stream and ignoring the vision. Making latents appear visual is therefore insufficient; the objective must force predictions to depend on them.

Motivated by this, we introduce \textbf{AnchorSim} (\emph{Anchored Mental Simulation}) to train interleaved multimodal models to rely on their generated visual pathway. During training, AnchorSim contrasts a latent-aware forward pass with a latent-blind pass (where visual inputs and latents are masked). An anti-collapse objective pushes the latent-aware predictions away from the latent-blind distribution, with stop-gradient on the blind branch, forcing the model to encode consequential information in the visual states rather than relying solely on text. To stabilize these continuous latents, this contrastive pressure is combined with distillation from an exponential moving average teacher. At inference, the auxiliary blind branch is discarded, and the model generates standard interleaved trajectories.

We evaluate AnchorSim across nine visual reasoning benchmarks against state-of-the-art latent~\cite{covt, skila, mirage, lvr} and RL-enhanced~\cite{visionr1, pixelreasoner} MLLMs. AnchorSim achieves the best average, improves over the base model by $13.5$ points, and ranks first on seven of nine benchmarks.
Crucially, diagnostic evaluations and ablations confirm these gains stem from genuine visual grounding: perturbing the visual latents now meaningfully impacts predictions. These findings demonstrate that AnchorSim does not merely produce visual states in form, but mitigates collapse by making them functional components of multimodal reasoning. Our work provides the community with both a critical diagnostic framework to identify simulation collapse and a principled blueprint for building models that genuinely reason across modalities. 

\section{Related Work}
\custompar{Latent Reasoning}
Latent reasoning studies how models can perform intermediate computation in hidden or continuous states rather than expressing every step as natural language. In language models, this idea appears in approaches that introduce thinking tokens, recurrent latent states, or additional test-time computation before decoding the final answer~\cite{coconut, goyal2023think, shen2025codi}. These methods aim to increase reasoning capacity without requiring long explicit chains of text~\cite{wei2022chain, wang2022self}. Recent multimodal work extends this idea to visual reasoning by interleaving text with visual thought tokens or latent image-conditioned states~\cite{mirage, lvr, covt, skila, monet, ilvr, mull}. Systems such as Mirage~\cite{mirage} suggest that multimodal models can reason through internal visual representations without always rendering pixels. Our work follows this broader direction, but focuses on a different question: not how to insert latent thoughts into the trajectory, but whether those thoughts are actually used. We show that existing latent visual reasoning models can emit visual latents while leaving the final prediction largely insensitive to them, a failure mode we call \emph{Mental Simulation Collapse}. AnchorSim contributes to latent reasoning by adding an explicit anti-collapse training signal: it contrasts predictions made with the visual pathway against predictions made when that pathway is removed, encouraging latent thoughts to become causally relevant rather than merely present.

\custompar{Visual Reasoning}
Recent progress in vision-language models~\cite{Qwen2.5-VL, li2024llava, chen2024internvl, hurst2024gpt} has substantially improved visual perception, but robust reasoning over visual inputs remains a persistent challenge. Motivated by the success of chain-of-thought prompting~\cite{wei2022chain} in language models, many works extend textual reasoning strategies to multimodal tasks, asking models to verbalize intermediate steps before producing an answer~\cite{visionr1, feng2025videor1, pixelreasoner}. While effective in some settings, purely textual reasoning can drift away from the visual evidence~\cite{luo2026thinking} and be a poor fit for problems that require tracking spatial structure~\cite{cvbench, sat}, object transformations~\cite{clevrer, osc}, or intermediate visual states~\cite{coin, xue2025progress}. This limitation has led to a growing body of work on visual forms of reasoning, including methods that generate explicit image-like thoughts~\cite{xu2025visualplanningletsthink, tong2025thinking, wiedemer2025video, vbvr, zhang2025latent, wang2026demystifing}, use unified multimodal models to interleave text and visual outputs~\cite{wu2024vila, liu2025javisgpt, deng2025bagel, zebra, gu2025thinkmorph, mvot}, introduce latent reasoning tokens~\cite{mirage, monet, lvr, covt}, or call external tools such as drawing, cropping, detection, and code execution~\cite{zhang2025thinking, yang2025longvt, fan2025grit, r3, ou2025bridging, zheng2025deepeyes, hong2025deepeyesv2}. These approaches broaden the reasoning interface beyond language, but each comes with trade-offs: pixel-level generation can be expensive, tool-based reasoning depends on brittle predefined operations, and latent reasoning is efficient but may not guarantee that the visual states are actually used.

Our work contributes to this latent visual reasoning line by shifting the focus from producing visual thoughts to making them consequential. Rather than treating latent visual states as useful merely because they are interleaved with text or aligned to visual embeddings, we directly test whether perturbing them changes the model's prediction. This reveals Mental Simulation Collapse, a failure mode in which models emit visual latents but route most answer-relevant computation through the linguistic stream. AnchorSim addresses this gap with a training objective that contrasts latent-aware and latent-blind predictions, explicitly making it suboptimal for the model to ignore its generated visual pathway. In doing so, our work complements recent efforts on spatial and multi-step visual reasoning benchmarks and training data, while providing a general anti-collapse mechanism for making latent visual reasoning functionally grounded rather than only architecturally present. Concurrent work, CapImagine~\cite{li2026imagination}, finds that perturbing visual latents barely changes benchmark accuracy and shifts reasoning to explicit text. In contrast, we apply matched perturbations to both the visual and textual streams, analyze why the training objective permits collapse, and keep the latent pathway while penalizing the model for ignoring it.
 
\section{Does the Mental Imagery Matter? Diagnosing Mental Simulation Collapse}
\label{sec:latent_collapse}

A recent line of works ~\cite{mirage, ilvr} propose that interleaving \emph{visual latent blocks} into an autoregressive reasoning trajectory can give multimodal models a form of mental imagery: dedicated image-conditioned hidden states that the model can consult while reasoning. The idea is compelling, but it leaves open a basic question. When a model emits a trajectory of the form:
\[
\langle \textcolor[HTML]{8A63B0}{\texttt{text}}_0, \textcolor[HTML]{2F9C95}{\texttt{vis}}_0, \textcolor[HTML]{8A63B0}{\texttt{text}}_1, \textcolor[HTML]{2F9C95}{\texttt{vis}}_1,
\dots, \textcolor[HTML]{8A63B0}{\texttt{text}}_n, \textcolor[HTML]{4E86B8}{\texttt{answer}}\rangle,
\]
do the visual latent blocks actually affect the answer, or does the model rely mostly on the surrounding language? We answer this question with a direct perturbation test and find that, across representative latent-reasoning systems, the generated visual blocks contribute substantially less to the model's own answer than the linguistic blocks do. We refer to this failure mode as
\textbf{Mental Simulation Collapse} (MSC): \textit{the model's trajectory exhibits the surface form of visual reasoning, yet the answer-relevant computation is carried almost entirely by the textual stream.}

 \begin{figure}[t]
    \centering
    \includegraphics[width=\linewidth]{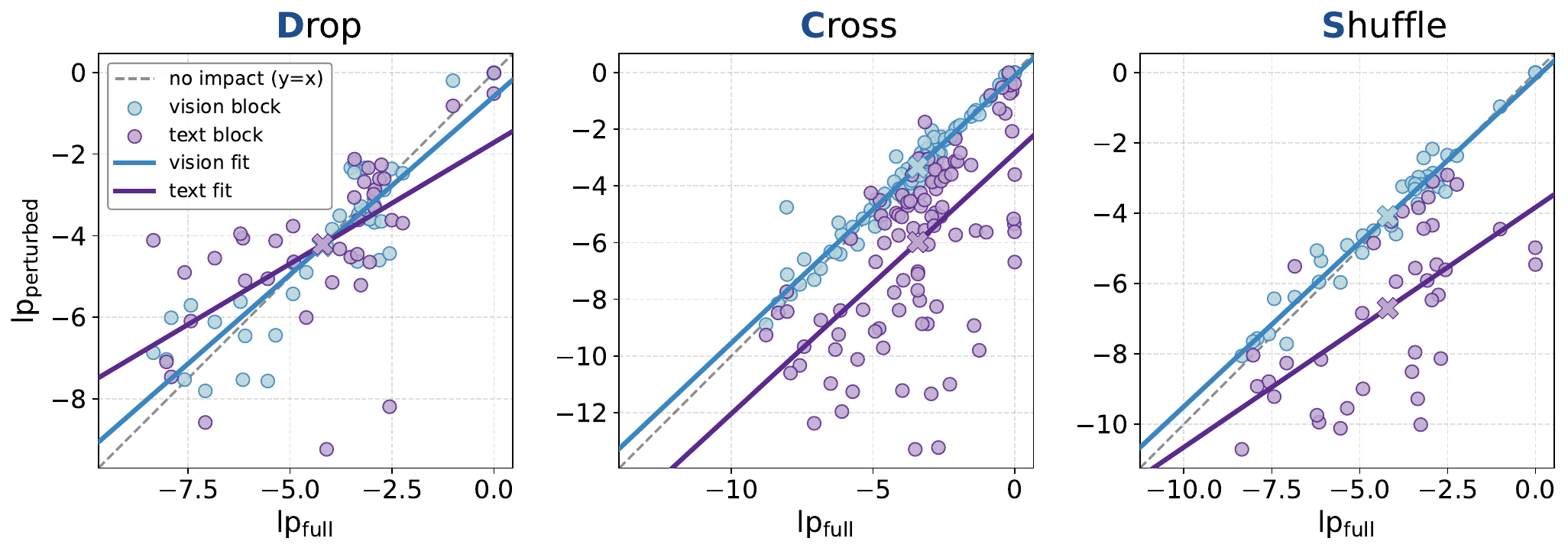}
\caption{
  \textbf{Per-sample vision vs.\ text perturbation impact for Mirage-7B.} 
  Each panel plots the perturbed final-answer log-probability ($y = \mathrm{lp}_{\mathrm{perturbed}}$) against the unperturbed baseline ($x = \mathrm{lp}_{\mathrm{full}}$) for the \emph{vision block} (blue) and \emph{reasoning text block} (purple). Points further below the $y{=}x$ diagonal (dashed) and lower ordinary-least-squares fit slopes (solid lines) indicate a larger performance drop. Across all conditions, the text fit has a lower slope than the vision fit, demonstrating that perturbing the linguistic chain degrades the answer more than perturbing the visual latents.
}
    \label{fig:mirage_scatter}
  \end{figure}

\subsection{Quantifying Latent Dependence: A Perturbation Probe}
\custompar{Setup} We start from four recent latent-reasoning MLLMs: \textsc{Mirage-7B}~\cite{mirage}, \textsc{SkiLa-7B}~\cite{skila}, \textsc{CoVT-7B}~\cite{covt}, and \textsc{Monet-7B}~\cite{monet}. For each model, we randomly select $100$ samples for a pivot study. The samples are drawn from CoMT~\cite{comt}, BLINK~\cite{blink}, V\textsuperscript{*}~\cite{vstar}, and MMVP~\cite{mmvp}+V\textsuperscript{*}~\cite{vstar}, respectively, for the four models. We use different benchmarks because, for some model--benchmark pairs, the model exhibits insufficient multi-hop reasoning behavior, defined as having fewer than two reasoning blocks.
The probing process consists of two steps: (i) \textbf{Trajectory capture:} We record the model's greedy-decoded trajectory along with the hidden states at every visual latent slot. (ii) \textbf{Teacher-forced scoring.} We feed the captured trajectory back through the model under teacher forcing and record $\mathrm{lp}_{\mathrm{full}}$, the log-probability assigned to the answer span.

To localize where the answer-relevant computation resides, we re-score the same answer span after perturbing the captured trajectory, applying perturbations independently on the vision side or the text side. Vision-side perturbations target only the visual latent hidden states, while text-side perturbations target only the natural-language reasoning segments; in both cases, the rest of the trajectory is left intact. We consider four conditions: (i) \texttt{Full} leaves the trajectory unchanged and serves as the
reference, (ii) \texttt{Drop} removes the block content, (iii) \texttt{Cross} replaces each block with the corresponding block from a donor sample, and (iv) \texttt{Shuffle} permutes the model's own blocks across positions, preserving real content but
moving it to the wrong locations. The answer span itself is never modified, so any shift in log-probability reflects the contribution of the perturbed block to the model's prediction. Implementation details of each perturbation are provided in Appendix~\ref{app:perturbation_details}.

\custompar{Vision Blocks Barely Move the Answer}
Figure~\ref{fig:vision_vs_text_all_models} summarizes the result. In most conditions, the blue vision bars under \texttt{Drop}, \texttt{Cross}, and \texttt{Shuffle} remain close to the
\texttt{Full} reference, while the purple text bars fall noticeably lower. The same pattern appears consistent across all four models and their corresponding benchmarks: removing, replacing, or permuting visual latent blocks costs the model little log-probability on its own answer, whereas applying the same perturbations to the linguistic reasoning blocks is typically more damaging. This asymmetry is striking because the vision-side perturbations directly target the latent states these architectures were designed to introduce. If the models were genuinely using their visual mental states as intermediate reasoning variables, their predictions should be highly sensitive to corruption of those states. Instead, the answer remains largely intact when the visual blocks are altered, suggesting that the model can bypass them and recover the answer from the textual stream.

\textbf{Per-sample Evidence: the Gap is Not Driven by Outliers.} To verify that the aggregate results in Figure~\ref{fig:vision_vs_text_all_models} are not driven by outliers, we examine \textsc{Mirage-7B} on a per-sample basis in Figure~\ref{fig:mirage_scatter}. Each panel plots $\mathrm{lp}_{\mathrm{perturbed}}$ against $\mathrm{lp}_{\mathrm{full}}$ for a single perturbation condition, with the dashed $y{=}x$ diagonal marking no impact and points farther below the diagonal indicating larger degradation. For each category, we fit an ordinary least-squares regression line to summarize the overall trend. The per-sample analysis reinforces the aggregate finding in Figure~\ref{fig:vision_vs_text_all_models}. For every perturbation condition, the text fit has a lower slope than the vision fit, indicating that the typical example loses more answer log-probability when its text block is perturbed than when its visual latent block is perturbed. The vision slopes ($0.87$ for \texttt{Drop}, $0.94$ for \texttt{Cross}, and $0.93$ for \texttt{Shuffle}) are closer to the no-impact slope of $1$ than the corresponding text slopes ($0.59$, $0.92$, and $0.68$). The gap is largest under \texttt{Drop} and \texttt{Shuffle}, the two conditions that most directly disrupt the structure of the reasoning trajectory. The asymmetry is therefore not driven by a small number of outliers: at the level of individual trajectories, the visual latent content is largely inert.

Taken together, these observations show that the visual latent blocks emitted by current latent-reasoning models behave less like load-bearing intermediate computations and more like decorative positions. Their hidden states can be dropped, swapped with another sample's, or shuffled out of order with only minor consequence for the model's own answer, whereas comparable perturbations to the natural-language reasoning chain produce much larger drops. This asymmetry echoes prior findings on text-only reasoning, where the \emph{structure} of the reasoning chain matters~\citep{li2025llms} and the reasoning ability can be improved even with invalid individual content~\citep{wang2023towards}. By the same diagnostic, the natural-language stream behaves like a genuine reasoning chain, and the visual latent stream does not. Although these architectures are capable of routing information through dedicated visual blocks, they have learned to channel the answer-relevant signal through language instead. MSC does not imply that these models gain nothing from latent blocks: they can still help through additional serial computation, a more structured reasoning trajectory, and changed attention to the input image, rather than through the content of the latents (Appendix~\ref{app:msc_improve}).

\subsection{Why Latent Visual Reasoning Collapses}
\label{sec:why_collapse}
The previous paragraphs demonstrated consistent evidence of Mental Simulation Collapse (MSC) across different latent block designs (token-pad, sketch-token, hidden-state, and multi-encoder variants), multiple benchmarks, and a range of trajectory lengths, suggesting that the problem is not a peculiarity of any single architecture, dataset, or decoding format. Instead, it points to a fundamental training-level failure: current objectives permit models to generate visual latents without making future predictions causally depend on them.

Latent-reasoning methods~\cite{mirage, ilvr} typically follow a two-stage training recipe. Let $z_\theta = G_\theta(q, \mathbf{v}, t_{\mathrm{pre}})$ denote the visual latent generated from query $q$, image $\mathbf{v}$, and textual prefix $t_{\mathrm{pre}}$. The decoder then predicts the answer through $p_\theta(y \mid q, \mathbf{v}, t_{\mathrm{pre}}, z_\theta)$. Training proceeds with
\begin{equation}
\mathcal{L}_{\mathrm{Stage1}}
=
\ell_{\mathrm{align}}(z_\theta, \hat z)
+
\gamma \mathcal{L}_{\mathrm{text}},
\qquad
\mathcal{L}_{\mathrm{Stage2}}
=
\mathcal{L}_{\mathrm{text}}
=
-\log p_\theta(y \mid q, \mathbf{v}, t_{\mathrm{pre}}, z_\theta).
\end{equation}

Stage 1 aligns $z_\theta$ to a helper-image embedding $\hat z$, while Stage 2 removes the alignment loss and optimizes only the text likelihood. The implicit assumption is that Stage 1 establishes visual grounding and Stage 2 preserves it while adapting the latent for reasoning. This objective leaves the latent pathway underdetermined at both stages: Stage 2 cannot distinguish latent-dependent reasoning from latent-independent shortcuts, while Stage 1 constrains how the latent looks, not whether it is used.

\paragraph{Text-based supervision does not identify latent dependence.}

In both stages, the text loss observes only the output distribution, not the causal pathway producing it. Consequently, latent usage is not identifiable under text-only supervision. In particular, for any latent-dependent solution, a sufficiently expressive decoder can achieve nearly identical likelihood while ignoring the latent:
\begin{equation}
p_{\theta'}(y \mid q, \mathbf{v}, t, z)
\approx
p_\theta(y \mid q, \mathbf{v}, t, G_\theta(q, \mathbf{v}, t))
\qquad \forall z.
\label{eq:unidentifiable}
\end{equation}

Thus, the objective cannot distinguish whether the latent acts as a visual workspace, a non-visual scratchpad, or an unused placeholder. Moreover, collapse is stable: once the decoder becomes invariant to $z$, gradients through the latent pathway become weak or uninformative, removing the signal needed to maintain meaningful latent computation. 

From an information-theoretic perspective~\cite{tishby2000information}, collapse corresponds to $I_\theta(y; z \mid q, \mathbf{v}, t) \approx 0$. The conditional mutual information $I_\theta(y; z \mid q, \mathbf{v}, t)$ measures how much the latent $z$ tells us about the output $y$ beyond what is already determined by the query $q$, image $\mathbf{v}$, and text prefix $t$; it equals zero precisely when $y$ and $z$ are conditionally independent given $(q, \mathbf{v}, t)$, i.e., when $p_\theta(y \mid q, \mathbf{v}, t, z) = p_\theta(y \mid q, \mathbf{v}, t)$. This is exactly the collapse condition: when the output distribution is invariant to $z$, the latent contributes no useful information, precisely the pattern observed in the probe experiments.

Such unidentifiability is not resolved by the visual alignment in the first stage. The alignment loss ($\ell_{\mathrm{align}}(z_\theta, \hat z)$) compares $z_\theta$ to a target embedding $\hat z$, and never inspects whether downstream computation uses the latent. Even when alignment is fully achieved and $z_\theta \approx \hat z$, the decoder may still satisfy $p_\theta(y \mid q, \mathbf{v}, t, z_\theta) \approx p_\theta(y \mid q, \mathbf{v}, t)$, leaving the latent visually plausible yet functionally inert. Adding visual supervision constrains how the latent \emph{looks} but not how it is \emph{used}, so the same condition $I_\theta(y; z \mid q, \mathbf{v}, t) \approx 0$ persists.

\begin{figure}[t]
    \centering
    \includegraphics[width=\linewidth]{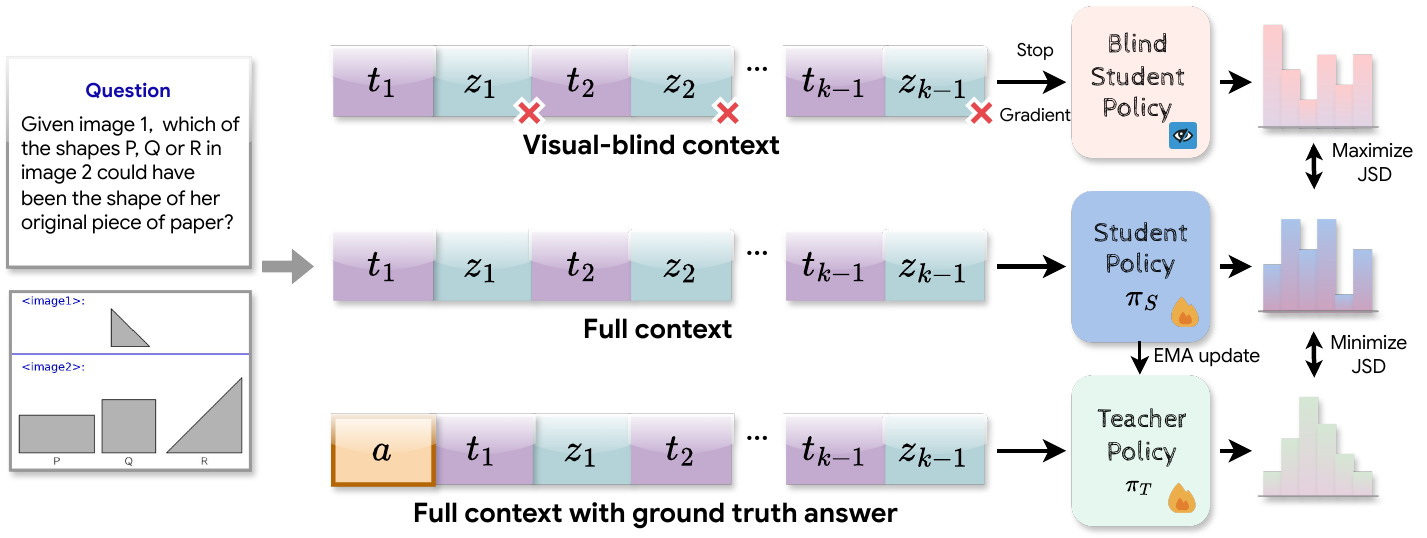}
\caption{
Overview of AnchorSim. Given an initial image and query, the model produces an interleaved reasoning trajectory consisting of textual thoughts $t_k$ and intermediate visual simulations $z_k$. AnchorSim trains the student policy with two complementary signals. First, a visual-blind branch masks the previous generated visual state $z_{k-1}$ and is stopped from receiving gradients; the student is encouraged to diverge from this blind prediction, preventing ``lazy text'' reasoning that ignores visual latents. Second, a simulation-anchored teacher, updated as an EMA of the student and conditioned on the ground-truth answer, provides reliable reasoning targets under the full context. The student is optimized to minimize the JSD to the teacher while maximizing the JSD from its visual-blind counterpart, thereby anchoring textual reasoning to the generated visual simulations.
}    \label{fig:framework}
\end{figure}

\section{Anchored Mental Simulation (AnchorSim)}
\label{sec:method}
The empirical probe and theoretical analysis point to the same key takeaway: mental simulation fails not because current models lack visual latent states, but because training gives the decoder no reason to make its future predictions depend on those states. The perturbation results show that visual blocks are largely non-causal in practice, while the objective-level analysis explains why this behavior is allowed: text likelihood can be optimized through latent-independent shortcuts. Therefore, an effective method must do more than generate visually plausible intermediate states; it must explicitly make the textual reasoning distribution sensitive to the presence and content of those states. This motivates our \textbf{AnchorSim} (Anchored Mental Simulation), which directly targets the collapse mechanism by contrasting full-context predictions against visual-blind predictions and distilling reasoning that remains anchored to the generated visual simulations.

AnchorSim is a lightweight training algorithm that encourages causal dependence between the model's reasoning and its visual latents. It is built on two core components: (i) \emph{Contrastive Latent Anchoring}, which enforces a measurable difference between the model's predictions with and without visual content; and (ii) \emph{Anchored Reasoning Distillation}, which supplies high-quality reasoning that coherently bridges each latent to the final answer. Figure~\ref{fig:framework} illustrates the overall framework.

Given an initial visual input $\mathbf{v}$ and a textual query $q$, the model produces an interleaved reasoning trajectory $\mathcal{T} = (t_1, z_1, \dots, t_K, z_K, a)$, where $t_k$ represents a text reasoning chunk, $z_k$ represents a generated intermediate visual state (mental simulation), and $a$ is the final answer. When predicting the $k$-th text chunk, the \emph{full context} is defined as $C_{<k} = (q, \mathbf{v}, t_1, z_1, \dots, t_{k-1}, z_{k-1})$. We further define an \emph{visual-blind context} $\tilde{C}_{<k}$, wherein all previous visual state $\{z_1, z_2, \dots, z_{k-1}\}$ is masked from the attention mechanism: $\tilde{C}_{<k} = C_{<k} \setminus \{z_1, z_2, \dots, z_{k-1}\}$.

\subsection{Contrastive Latent Anchoring}
The diagnosis in Section~\ref{sec:why_collapse} identified the core problem: standard next-token training does not require the model's predictions to depend on the visual latent stream, so the model is free to ignore it whenever language priors alone yield a plausible continuation. Contrastive Latent Anchoring closes this loophole directly. For every reasoning step, we form two next-token distributions from the student policy $\pi_S$: one conditioned on the full context $C_{<k}$, and one conditioned on the visual-blind context $\tilde{C}_{<k}$ in which all preceding latents $\{z_1, \ldots, z_{k-1}\}$ have been masked. We then penalize agreement between the two distributions, since it implies that the latent stream contributed nothing:
\begin{equation}
\mathcal{L}_{\text{anchor}}(\theta) = -\,\mathbb{E}_{\mathcal{T}, k}\!\left[ \frac{1}{|t_k|} \sum_{n=1}^{|t_k|} \mathrm{JSD}\!\Big(\pi_S(\cdot \mid C_{<k}, t_{k,<n}) \,\big\|\, \mathrm{sg}\!\big[\pi_S(\cdot \mid \tilde{C}_{<k}, t_{k,<n})\big]\Big) \right],
\label{eq:anchor}
\end{equation}
where $\pi_S$ is the student policy, $\mathrm{JSD}$ denotes the Jensen--Shannon divergence~\cite{lin2002divergence} and $\mathrm{sg}[\cdot]$ stops gradients through the visual-blind branch. Minimizing $\mathcal{L}_{\text{anchor}}$ enforces a measurable gap between predictions with and without the visual stream. 

\subsection{Anchored Reasoning Distillation}
To ensure the text remains logically sound while anchored to the visual state, we distill high-quality reasoning pathways from a privileged teacher $\pi_T$. Rather than acting as a static target, the teacher shares the student's architecture and is parameterized by an exponential moving average (EMA) of the student's weights $\bar{\theta} \leftarrow \alpha \cdot  \bar{\theta} + (1-\alpha) \cdot \theta$, motivated by~\cite{moco}.

The teacher is conditioned on the ground-truth answer $a$, which is injected into its prompt as a privileged prefix~\cite{opsd}~\cite{dugad2026watching}. This access lets the teacher reliably interpret each generated latent $z_{k-1}$ and produce textual reasoning that coherently bridges the visual state to the correct answer. We distill this answer-anchored reasoning into the student via
\begin{equation}
    \mathcal{L}_{\text{distill}}(\theta) = \mathbb{E}_{\mathcal{T}, k}\!\left[ \frac{1}{|t_k|} \sum_{n=1}^{|t_k|} \mathrm{JSD}\!\Big(\pi_T(\cdot \mid C_{<k}, a, t_{k,<n}) \,\big\|\, \pi_S(\cdot \mid C_{<k}, t_{k,<n})\Big) \right],
\label{eq:distill}
\end{equation}
where the student and teacher consume the same student-generated latents $z_{k-1}$; the teacher's role is purely interpretive, supplying a learning target that uses the latents faithfully.

\subsection{The AnchorSim Training Objective}
\label{sec:method:unified}
We train the model end-to-end by minimizing the \emph{AnchorSim} loss, which is a weighted sum of the distillation and anchoring objectives from the previous subsections:
\begin{equation}
    \mathcal{L}_{\text{AnchorSim}}(\theta) = \mathcal{L}_{\text{distill}}(\theta) + \lambda \, \mathcal{L}_{\text{anchor}}(\theta),
\label{eq:anchorsim}
\end{equation}
where $\lambda > 0$ scales the anchoring penalty. Contrastive anchoring encourages the latents to carry answer-relevant, image-conditioned information that affects subsequent predictions, while answer-conditioned distillation teaches the model to use that content productively.
\section{Experiments}
\label{sec:exp}
\begin{table}[t]
\centering
\caption{Comparison with proprietary and open-source baselines across 9 public visual reasoning benchmarks. Best and second-best results are highlighted in \best{bold underlined} and \sbest{bold} style, respectively. We report BLINK with both the Jigsaw subset and overall average, CV-Bench with its 2D and 3D splits, and Avg as the average performance over 9 benchmarks, where BLINK uses its overall average and CV-Bench uses the mean of its 2D and 3D splits. $^{*}$Results are reported from prior work.}
\resizebox{\textwidth}{!}{%
\begin{tabular}{lcccccc}
\thickhline
\multirow{2}{*}{\textbf{Model}} 
& \multicolumn{2}{c}{\textbf{BLINK}} 
& \multirow{2}{*}{\textbf{V*}} 
& \multirow{2}{*}{\textbf{MMVP}} 
& \multirow{2}{*}{\textbf{EMMA}} 
& \multirow{2}{*}{\textbf{VisuLogic}} \\
\cline{2-3}
& \textbf{Jigsaw} & \textbf{Avg} & & & & \\
\hline
\rowcolor{LighterGray}
GPT-4o~\cite{hurst2024gpt}* & - & \best{60.0} & 66.0 & 70.7 & - & - \\
\rowcolor{LighterGray}
Claude-4-Sonnet~\cite{anthropic2025claude}* & - & 39.6 & 15.2 & - & - & - \\
Vision-r1~\cite{visionr1}* & 27.3 & 51.0 & 80.1 & 72.6 & 24.3 & 15.2 \\
PixelReasoner~\cite{pixelreasoner}* & 52.7 & - & 80.6 & - & 23.5 & 23.4 \\
LVR~\cite{lvr}* & 52.7 & 52.5 & 81.7 & 72.0 & - & - \\
CoVT~\cite{covt} & 62.7 & 56.0 & 78.0 & \sbest{77.3} & 22.7 & 24.6 \\
SkiLa~\cite{skila} & 45.3 & 53.0 & \best{84.3} & 75.3 & 21.5 & 24.1 \\
\hline
\textsc{Qwen-2.5-VL-7B}~\cite{Qwen2.5-VL} & 61.3 & 55.5 & 77.5 & 74.0 & 24.3 & 25.6 \\
\rowcolor{LightPurple}
\textsc{Qwen-2.5-VL-7B} (interleaved)
& \sbest{64.7}~{\textcolor{ForestGreen}{\footnotesize (+3.4)}} 
& 58.0~{\textcolor{ForestGreen}{\footnotesize (+2.5)}} 
& 79.1~{\textcolor{ForestGreen}{\footnotesize (+1.6)}} 
& 75.7~{\textcolor{ForestGreen}{\footnotesize (+1.7)}} 
& \sbest{25.5}~{\textcolor{ForestGreen}{\footnotesize (+1.2)}} 
& \sbest{25.8}~{\textcolor{ForestGreen}{\footnotesize (+0.2)}} \\
\rowcolor{LightPurple}
\textsc{AnchorSim-7B} (Ours)
& \best{66.0}~{\textcolor{ForestGreen}{\footnotesize (+4.7)}} 
& \sbest{59.4}~{\textcolor{ForestGreen}{\footnotesize (+3.9)}} 
& \sbest{82.7}~{\textcolor{ForestGreen}{\footnotesize (+5.2)}} 
& \best{78.3}~{\textcolor{ForestGreen}{\footnotesize (+4.3)}} 
& \best{26.0}~{\textcolor{ForestGreen}{\footnotesize (+1.7)}} 
& \best{27.5}~{\textcolor{ForestGreen}{\footnotesize (+1.9)}} \\
\thickhline
\multirow{2}{*}{\textbf{Model}} 
& \multicolumn{2}{c}{\textbf{CV-Bench}} 
& \multirow{2}{*}{\textbf{SAT-R}} 
& \multirow{2}{*}{\textbf{VSP}} 
& \multirow{2}{*}{\textbf{CoMT}} 
& \multirow{2}{*}{\textbf{Avg}} \\
\cline{2-3}
& \textbf{2D} & \textbf{3D} & & & & \\
\hline
Vision-r1~\cite{visionr1}* & 8.0 & 37.6 & - & - & - & - \\
CoVT~\cite{covt} & 76.0 & \sbest{84.7} & 64.7 & 6.0 & 27.8 & 48.6 \\
SkiLa~\cite{skila} & \sbest{76.7} & 74.4 & 56.7 & 4.3 & 32.3 & 47.5 \\
\hline
\textsc{Qwen-2.5-VL-7B}~\cite{Qwen2.5-VL} & 76.6 & 73.3 & 59.7 & 9.3 & 39.5 & 48.9 \\
\rowcolor{LightPurple}
\textsc{Qwen-2.5-VL-7B} (interleaved)
& \sbest{76.7}~{\textcolor{ForestGreen}{\footnotesize (+0.1)}} 
& 77.3~{\textcolor{ForestGreen}{\footnotesize (+4.0)}} 
& \sbest{65.3}~{\textcolor{ForestGreen}{\footnotesize (+5.6)}} 
& \sbest{74.8}~{\textcolor{ForestGreen}{\footnotesize (+65.5)}} 
& \sbest{56.3}~{\textcolor{ForestGreen}{\footnotesize (+16.8)}} 
& \sbest{59.7}~{\textcolor{ForestGreen}{\footnotesize (+10.8)}} \\
\rowcolor{LightPurple}
\textsc{AnchorSim-7B} (Ours)
& \best{80.4}~{\textcolor{ForestGreen}{\footnotesize (+3.8)}} 
& \best{86.4}~{\textcolor{ForestGreen}{\footnotesize (+13.1)}} 
& \best{68.7}~{\textcolor{ForestGreen}{\footnotesize (+9.0)}} 
& \best{78.0}~{\textcolor{ForestGreen}{\footnotesize (+68.7)}} 
& \best{58.0}~{\textcolor{ForestGreen}{\footnotesize (+18.5)}} 
& \best{62.4}~{\textcolor{ForestGreen}{\footnotesize (+13.5)}} \\
\thickhline
\end{tabular}
}
\vspace{-5pt}
\label{tab:visual_reasoning_results}
\end{table}

\custompar{Training Strategy}
We adopt \textsc{Qwen2.5-VL-7B}~\cite{Qwen2.5-VL} as our base model, following common practice in prior  works~\cite{mirage, covt, lvr}. We post-train it in two stages: \textbf{Stage 1} warms up the model for multimodal interleaved reasoning following Mirage~\cite{mirage}; \textbf{Stage 2} further post-trains the Stage 1 checkpoint with our proposed AnchorSim objective. Both stages use the same training mixture, drawn from Zebra-CoT~\cite{zebra}, SAT~\cite{sat}, CoMT~\cite{comt}, and VSP~\cite{vsp}. Since SAT does not provide interleaved reasoning chains, we supervise only its final-answer tokens during training. Detailed dataset statistics are reported in the Appendix. We refer to the resulting model as \textsc{AnchorSim-7B}.

\custompar{Benchmarks}
We evaluate on 9 public visual reasoning benchmarks that may benefit from latent-space mental imagery, organized into four families:
(i) \textit{perception and fine-grained visual search}: BLINK~\citep{blink}, MMVP~\cite{mmvp}, and V\textsuperscript{*}~\cite{vstar};
(ii) \textit{2D and 3D spatial reasoning}: CV-Bench-2D, CV-Bench-3D~\cite{cvbench}, and SAT-Real~\cite{sat};
(iii) \textit{abstract and multimodal STEM reasoning}: EMMA~\cite{emma} and VisuLogic~\cite{visulogic}; and
(iv) \textit{visual planning and step-wise manipulation}: VSP~\cite{vsp} and CoMT~\cite{comt}.
Together, these benchmarks cover low-level perception, spatial and depth reasoning, diagram-grounded STEM problem solving, abstract pattern induction, planning, and multi-step visual-state tracking.

\custompar{Implementation Details}
We use an EMA teacher with decay rate $\alpha = 0.99$. The latent block size is set to $8$ tokens by default. Stage 1 is trained for $3$ epochs, after which we follow Mirage~\cite{mirage} and fine-tune for $1$ additional epoch with text-only supervision. Stage 2 is initialized from the Stage 1 checkpoint and trained for $1$ further epoch with AnchorSim. All training is performed on $8$ NVIDIA H200 GPUs with a per-GPU batch size of $4$. 

\custompar{Baselines}
We compare \textsc{AnchorSim-7B} against both proprietary and open-source MLLMs. For proprietary systems, we include GPT-4o~\cite{hurst2024gpt} and Claude-4-Sonnet~\cite{anthropic2025claude}. For open-source baselines, we compare with two families: (i) reinforcement-learning-enhanced MLLMs, namely Vision-R1~\cite{visionr1} and PixelReasoner~\cite{pixelreasoner}; and (ii) recent latent-space reasoning models with publicly released checkpoints, including LVR~\cite{lvr}, CoVT~\cite{covt}, and SkiLa~\cite{skila}. For CoVT, we use the released \textsc{CoVT-7B-seg\_depth\_dino} checkpoint throughout. We also compare against our base model \textsc{Qwen2.5-VL-7B}~\cite{Qwen2.5-VL} and the Stage 1 model trained with Mirage~\cite{mirage}. Together, these baselines span a broad spectrum of contemporary vision-language systems: from large commercially deployed models to lightweight community releases---ensuring that our evaluation is both representative of the strongest available competitors. Further details on the evaluation setup are provided in Appendix~\ref{app:implementation_details}.

\subsection{Main Results}
\custompar{Anchor-Based Simulation Improves Visual Reasoning} Table~\ref{tab:visual_reasoning_results} compares \textsc{AnchorSim-7B} with proprietary models and recent open-source visual reasoning baselines across the 9 benchmarks spanning perception, spatial reasoning, abstract logic, STEM reasoning, planning, and step-wise manipulation. \textsc{AnchorSim-7B} achieves the strongest overall performance, attaining the best average score of $62.4\%$ and outperforming the \textsc{Qwen2.5-VL-7B} base by $13.5$ points and the Stage 1 interleaved model by $2.7$ points. The gains are consistent across reasoning categories: \textsc{AnchorSim-7B} ranks first on all benchmarks except BLINK and V\textsuperscript{*}, where it ranks second. This demonstrates strong generalization beyond any single benchmark family. The large improvements on VSP and CoMT in particular suggest that AnchorSim substantially benefits tasks requiring multi-step visual state tracking and planning. Compared with the Stage 1 Mirage model, \textsc{AnchorSim-7B} further improves performance on nearly all benchmarks, indicating that the proposed Stage 2 objective provides complementary gains by strengthening grounded latent-space visual reasoning. Figure~\ref{fig:qual_main} compares \textsc{AnchorSim-7B} and \textsc{SkiLa-7B} on SAT-R, where \textsc{AnchorSim-7B} predicts the correct answer and \textsc{SkiLa-7B} does not. Additional examples are provided in Appendix~\ref{app:qualitative_examples}.

\begin{figure}[t]
    \centering
    \includegraphics[width=\linewidth]{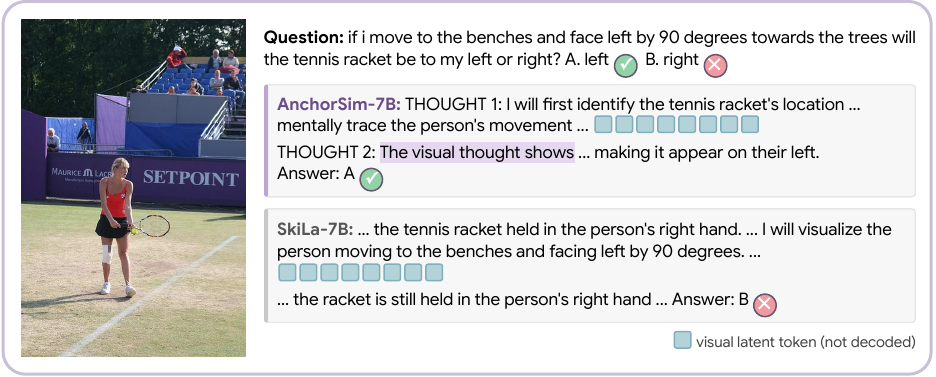}
    \caption{\textbf{Qualitative example on SAT-R.} \textsc{AnchorSim-7B} achieves better visual--text interleaving and correctly predicts the answer. Blue squares denote visual latent tokens.}
    \label{fig:qual_main}
\end{figure}

\subsection{Does AnchorSim Mitigate MSC?}
\label{sec:mitigate}

\custompar{Teacher-Forced Perturbation}
We apply the same perturbation analysis as in Section~\ref{sec:latent_collapse} to Qwen-interleaved and \textsc{AnchorSim}. The base \textsc{Qwen2.5-VL-7B} does not generate visual latents, so the same intervention cannot be applied to it. As shown in Figure~\ref{fig:perturbation_anchorsim}(a), \textsc{AnchorSim} makes visual perturbations more damaging while sensitivity to the text stream remains similar, reducing the gap $S_{\mathrm{vision}} - S_{\mathrm{text}}$ by $61\%$ under \texttt{Drop} and $44\%$ under \texttt{Shuffle}. Under \texttt{Cross}, the near-uniform downward shift is largely absorbed by the regression intercept, so slopes stay close to $1$ for both models. These results indicate that \textsc{AnchorSim} mitigates MSC. We next test whether this holds when the model regenerates its trajectory.

\begin{figure}[!t]
    \centering
    \includegraphics[width=\linewidth]{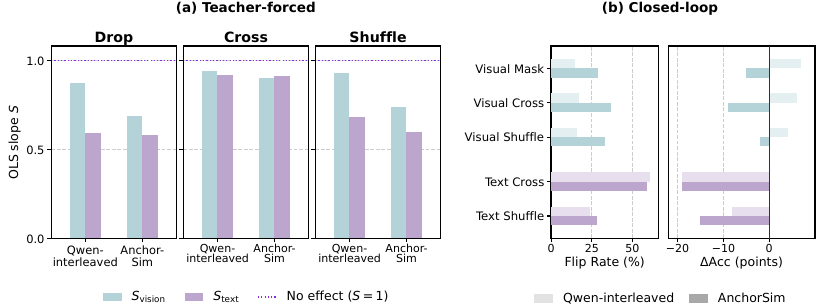}
    \caption{Perturbation analysis. (a) OLS slope $S$ between perturbed and unperturbed answer log-probabilities for Qwen-interleaved and \textsc{AnchorSim}, where $S \approx 1$ indicates little effect. (b) Closed-loop analysis, where the model regenerates the rest of the trajectory after the perturbation.}
    \label{fig:perturbation_anchorsim}
\end{figure}

\custompar{Closed-Loop Perturbation}
The analysis above holds the downstream reasoning fixed and measures direct answer dependence. To capture effects that propagate through subsequent reasoning, we perturb either the first visual latent block or the reasoning-text segment immediately following it in greedy rollouts of Qwen-interleaved and \textsc{AnchorSim}, and then regenerate all subsequent text, visual latent blocks, and the final answer. For the visual stream, we replace \texttt{Drop} with \texttt{Mask}, which prevents subsequent tokens from attending to the selected latent content, because after a literal deletion the model can immediately generate a replacement block. We report Flip Rate, the percentage of perturbations that change the final answer, and the change in accuracy relative to each model's unperturbed accuracy ($52\%$ for Qwen-interleaved and $60\%$ for \textsc{AnchorSim}). As shown in Figure~\ref{fig:perturbation_anchorsim}(b), Qwen-interleaved flips only $15$--$17\%$ of answers under visual perturbations and its accuracy does not decrease, whereas for \textsc{AnchorSim}, perturbing the latents flips $29$--$37\%$ of answers and reduces accuracy by up to $9$ points. In contrast, flip rates under text perturbations remain similar across the two models. The low visual sensitivity of the baseline is therefore not an artifact of teacher forcing. Together, these results show that \textsc{AnchorSim} makes visual latents consequential both for the answer and for the regenerated reasoning, mitigating MSC.

\subsection{Ablation Study}
\label{sec:ablation}

\noindent
\begin{minipage}[t]{0.64\linewidth}
    \vspace{0pt}
    \custompar{Effectiveness of Each Loss Term}
    We first ablate the contribution of each loss term in our objective. Compared with the \textsc{Qwen2.5-VL-7B} base model, training with the visual-distillation loss $\mathcal{L}_{\mathrm{distill}}$ consistently improves performance across all three benchmarks, yielding noticeable gains on BLINK, V$^{*}$, and CV-Bench. Further adding the contrastive loss $\mathcal{L}_{\mathrm{anchor}}$ boosts accuracy on every benchmark, with the largest marginal gain observed on V$^{*}$, where $\mathcal{L}_{\mathrm{anchor}}$ adds $+4.2$ points over the distillation-only variant. These results indicate that the two losses are complementary: distillation preserves task-relevant visual information, while the contrastive learning term further sharpens the learned representations by forcing the model to use them.
\end{minipage}
\hfill
\begin{minipage}[t]{0.33\linewidth}
    \vspace{0pt}
    \centering
    \includegraphics[width=\linewidth]{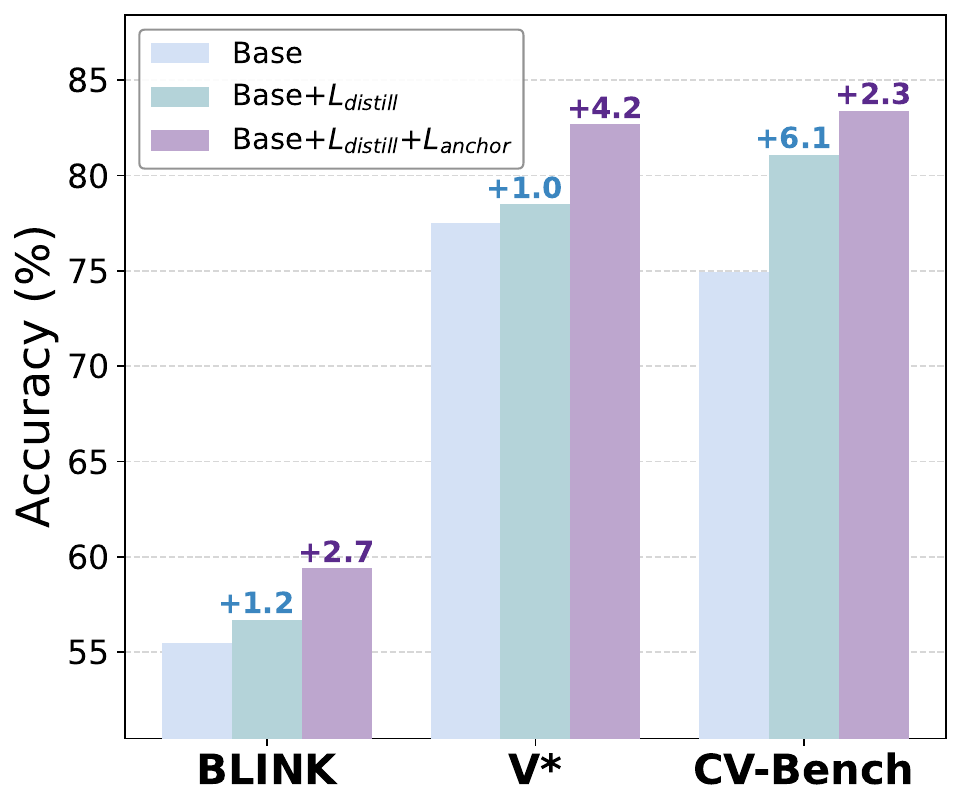}
    \captionof{figure}{\textbf{Loss ablation.}}
    \label{fig:ablation_contrastive}
\end{minipage}
\par\vspace{5pt}

\noindent
\begin{minipage}[t]{0.64\linewidth}
    \vspace{0pt}
    \custompar{Scalability with Number of Latent Tokens}
    We next study how performance scales with the number of visual latent tokens emitted per reasoning block. Introducing latent tokens improves over the no-latent Qwen base for every tested block size and benchmark, indicating that compact latent visual representations are beneficial even when only a small number of tokens are used. The trend, however, is not monotonic: BLINK and CV-Bench improve as the block size increases up to $8$ tokens but degrade when further increased to $16$, with CV-Bench dropping most noticeably. In contrast, V$^{*}$ continues to show a small improvement from $8$ to $16$ tokens. We therefore adopt $8$ latent tokens as the default setting, due to the best overall balance.
\end{minipage}
\hfill
\begin{minipage}[t]{0.33\linewidth}
    \vspace{0pt}
    \centering
    \includegraphics[width=\linewidth]{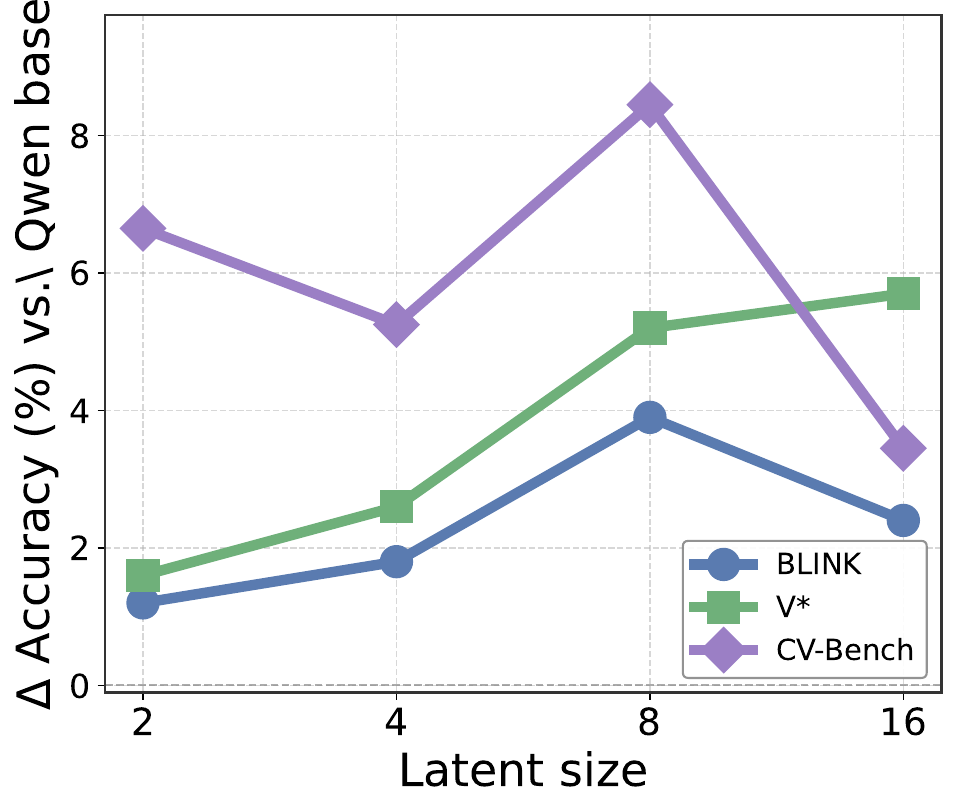}
    \captionof{figure}{\textbf{Effect of latent size.}}
    \label{fig:latent_size}
\end{minipage}
\par\vspace{5pt}

\section{Conclusions and Limitations}
We identify \emph{Mental Simulation Collapse} (MSC), where latent-reasoning multimodal models produce visual states that resemble mental simulation but barely affect final predictions, while answer-relevant computation is routed through text. To address this, we propose \textbf{AnchorSim} (\textbf{Anchor}ed Mental \textbf{Sim}ulation), a contrastive distillation framework that trains models to rely on generated visual latents instead of bypassing them through text-only shortcuts. Our results show that anchoring visual latents to model behavior improves visual reasoning and mitigates collapse by making latent simulations functional components of the reasoning process. \textsc{AnchorSim} is representation-agnostic: it does not require individual latent tokens to correspond to image crops or to be directly decodable, and we define their content operationally as answer-relevant, image-conditioned information.  AnchorSim is evaluated mainly at the 7B scale MLLM like \textsc{Qwen2.5-VL} on benchmark-style tasks, so future work should test its scalability to larger models and real-world multimodal decision-making. More broadly, we see making internal visual representations functionally necessary, rather than merely present, as a step toward models that can genuinely simulate before they act.
}

\newpage
\clearpage
\paragraph{Acknowledgements} This research has been supported by computing support on the Vista GPU Cluster through the Center for Generative AI (CGAI) and the Texas Advanced Computing Center (TACC) at the University of Texas at Austin. UT Austin is supported by NSF AI Institute for Foundations of Machine Learning (IFML) 2019844, NSF Grants 2019844, 2112471, AF 1901292, CNS 2148141, Tripods CCF 1934932, Tripods 2217069, the Texas Advanced Computing Center (TACC) and research gifts by Western Digital, Wireless Networking and Communications Group (WNCG) Industrial Affiliates Program, UT Austin Machine Learning Lab (MLL), and Cisco.

We thank the anonymous reviewers for their constructive feedback.

{
    \small
    \bibliographystyle{plain}
    \bibliography{main}
}

\newpage
\appendix
\appendix
\tableofcontents

\section{Additional Details for the Vision vs.\ Text Perturbation Study}
\label{app:perturbation_details}

This section provides implementation details for the perturbation probe introduced in Section~\ref{sec:latent_collapse}. The goal of the probe is to measure whether generated visual latent blocks causally affect the model's own final-answer probability, relative to the surrounding linguistic reasoning trace.

\paragraph{Fair text-side perturbations.}
A naive comparison between vision-side and text-side \texttt{Drop} is unfair for multi-hop trajectories. A typical latent-reasoning trajectory contains an initial planning segment, alternating visual latent blocks and between-block reasoning text, and a final segment that often explicitly states the answer, such as \textit{``Therefore, the answer is B''}. Vision-side \texttt{Drop} removes only the latent hidden states while leaving the entire linguistic scaffold intact. In contrast, naively dropping all text would also remove the answer-stating tail, degrading performance for reasons unrelated to intermediate reasoning.

To make the comparison more balanced, for multi-hop models we define text-side \texttt{Drop} as removing only the \emph{between-block reasoning text}, while preserving both the initial planning segment and the final answer-stating tail. This mirrors the asymmetry of vision-side \texttt{Drop}, where only intermediate visual content is neutralized. Monet emits a single visual block and does not contain between-block text, so we use the naive full-text \texttt{Drop} variant for that model.

\paragraph{Filtering and perturbation conditions.}
For \texttt{Drop} and \texttt{Shuffle}, we restrict evaluation to trajectories with at least two visual blocks whenever the perturbation requires a meaningful within-trajectory permutation. This applies to Mirage, SkiLa, and CoVT. Monet emits one visual block per trajectory, so vision-side \texttt{Shuffle} is omitted.

We evaluate four models on their corresponding benchmarks with $100$ samples per model: Mirage-7B on CoMT, SkiLa-7B on BLINK multi-hop, CoVT-7B on V\textsuperscript{*}Bench, and Monet-7B on MMVP $+$ V\textsuperscript{*}Bench. The perturbation conditions are:
\begin{itemize}
    \item \textbf{Full}: the original captured trajectory is scored without modification.
    \item \textbf{Drop}: the target block content is removed or masked.
    \item \textbf{Cross}: the target block is replaced with the corresponding block from another sample.
    \item \textbf{Shuffle}: the model's own blocks are permuted across positions.
\end{itemize}
The answer span itself is never modified during scoring. Therefore, changes in final-answer log-probability reflect how much the perturbed component contributes to the model's prediction.

\paragraph{Mirage-7B per-sample statistics.}
For the Mirage-7B per-sample perturbation analysis in Figure~\ref{fig:mirage_scatter}, \texttt{Drop} and \texttt{Shuffle} are restricted to multi-hop trajectories with $n_{\mathrm{latents}} \geq 2$ ($n=39$), while \texttt{Cross} uses all $100$ samples. The ordinary-least-squares fit slopes are:
\[
\begin{aligned}
\texttt{Drop}:& \quad \text{vision}=0.87,\quad \text{text}=0.59,\\
\texttt{Cross}:& \quad \text{vision}=0.94,\quad \text{text}=0.92,\\
\texttt{Shuffle}:& \quad \text{vision}=0.93,\quad \text{text}=0.68.
\end{aligned}
\]
Across all three perturbations, the text fit has a lower slope than the vision fit, indicating that the linguistic trace carries more answer-relevant signal than the generated visual latents.

\section{Benchmark Details}
\label{app:benchmark_details}

We evaluate on public visual reasoning benchmarks spanning perception, spatial reasoning, abstract logic, STEM reasoning, planning, and step-wise manipulation.

\paragraph{Perception and fine-grained visual search.}
BLINK~\citep{blink} evaluates 14 core visual perception skills. We report both the overall average and the jigsaw subset, which requires visual completion and fine-grained part-whole reasoning. MMVP~\cite{mmvp} contains CLIP-blind image pairs designed to expose perceptual failures in modern vision-language models. V\textsuperscript{*}~\cite{vstar} evaluates fine-grained visual search in high-resolution images, where correct answers often depend on small or peripheral details.

\paragraph{2D and 3D spatial reasoning.}
CV-Bench~\cite{cvbench} contains 2,638 items derived from COCO, ADE20K, and Omni3D. Its 2D split tests spatial relations and counting, while its 3D split evaluates depth ordering and relative distance. We report these as CV-Bench-2D and CV-Bench-3D. SAT-Real~\cite{sat} is the real-image split of the Spatial Aptitude Test, containing 300 items across action-consequence, ego-movement, goal-aiming, object-movement, and perspective-taking categories.

\paragraph{Abstract and multimodal STEM reasoning.}
EMMA~\cite{emma} is a multimodal STEM benchmark covering chemistry, coding, mathematics, and physics. Solving EMMA problems often requires combining symbolic reasoning with iterative grounding in diagrams. VisuLogic~\cite{visulogic} contains 1,000 abstract visual-logic puzzles inspired by Raven's progressive matrices, testing pattern induction over shapes, colors, and transformations.

\paragraph{Visual planning and step-wise manipulation.}
VSP~\cite{vsp} is a 400-item maze-planning benchmark on $3 \times 3$ to $6 \times 6$ grids. We evaluate predictions by simulating the generated U/D/L/R plan against the maze and count success only if the agent reaches the goal without stepping into a hole. CoMT~\cite{comt} is a 400-item Chain-of-Manipulations benchmark that tests step-wise visual editing reasoning over creation, deletion, selection, and update operations. 

\section{Implementation Details}
\label{app:implementation_details}

\paragraph{Training and optimization.}
We use Qwen2.5-VL-7B as the base model. Stage 1 follows the interleaved multimodal warmup recipe used by Mirage, while Stage 2 applies the AnchorSim objective described in Section~\ref{sec:method}. We optimize with AdamW~\cite{adamw} using a peak learning rate of $1 \times 10^{-6}$. Unless otherwise stated, we use a latent block size of $8$ tokens and EMA decay rate $\alpha=0.99$.

\paragraph{Evaluation.}
All benchmark evaluations use the same prompting and decoding settings across comparable models. For planning-style tasks such as VSP, we evaluate by executing the predicted action sequence in the environment rather than relying on exact string matching. For free-form tasks such as CoMT, predictions are normalized before answer comparison.

\section{Additional Perturbation Results}
\label{app:additional_results}
\begin{figure}[t]
    \centering
    \includegraphics[width=\linewidth]{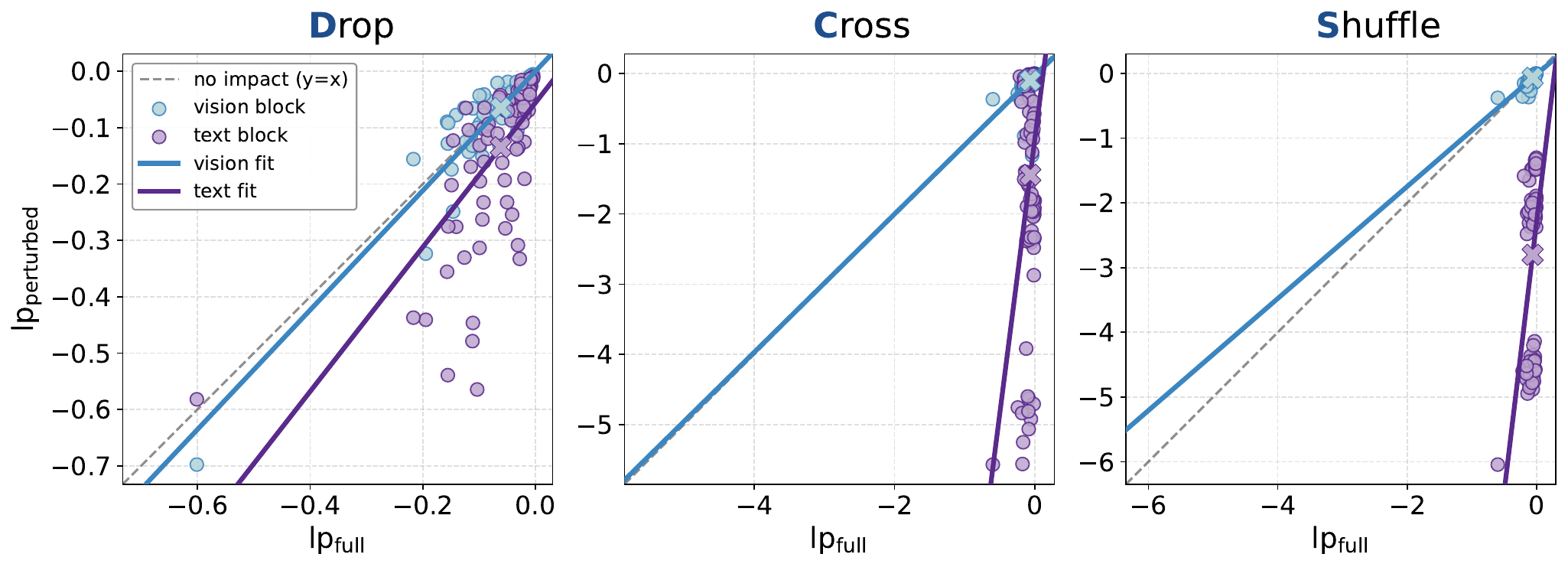}
    \caption{
    \textbf{Per-sample vision vs.\ text perturbation impact for SkiLa-7B on the multi-hop subset of BLINK.}
    The layout, axes, and color encoding match Figure~\ref{fig:mirage_scatter}. Each point compares the perturbed final-answer log-probability against the full-trajectory baseline. \texttt{Drop} and \texttt{Shuffle} are restricted to trajectories with $n_{\mathrm{sketches}}\geq2$ ($n=88$), while \texttt{Cross} uses all $100$ samples. Across conditions, text perturbations produce larger answer log-probability drops than sketch-hidden-state perturbations, indicating that the linguistic reasoning chain carries more answer-relevant computation than the visual sketch states.
    }
    \label{fig:skila_scatter}
\end{figure}

\begin{figure}[t]
    \centering
    \includegraphics[width=\linewidth]{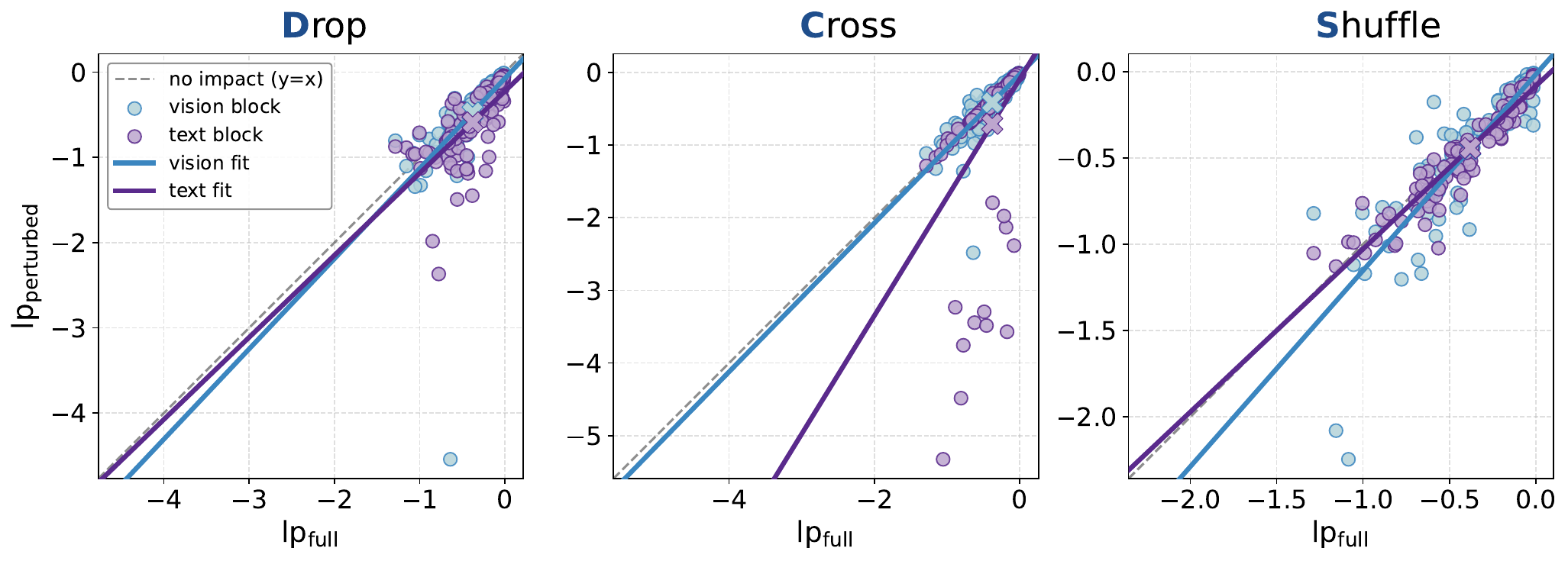}
    \caption{
    \textbf{Per-sample vision vs.\ text perturbation impact for CoVT-7B on V\textsuperscript{*}Bench.}
    CoVT emits a fixed three-encoder visual-block template per trajectory, so all perturbation conditions are evaluated on the full $100$-sample set. Vision-side perturbations leave many samples close to the no-impact diagonal, whereas text-side perturbations move a substantial subset below the diagonal. This shows that the model's final-answer probability is more sensitive to corrupted reasoning text than to corrupted SAM, depth, and DINO pad hidden states.
    }
    \label{fig:covt_scatter}
\end{figure}

\begin{figure}[!ht]
    \centering
    \includegraphics[width=0.7\linewidth]{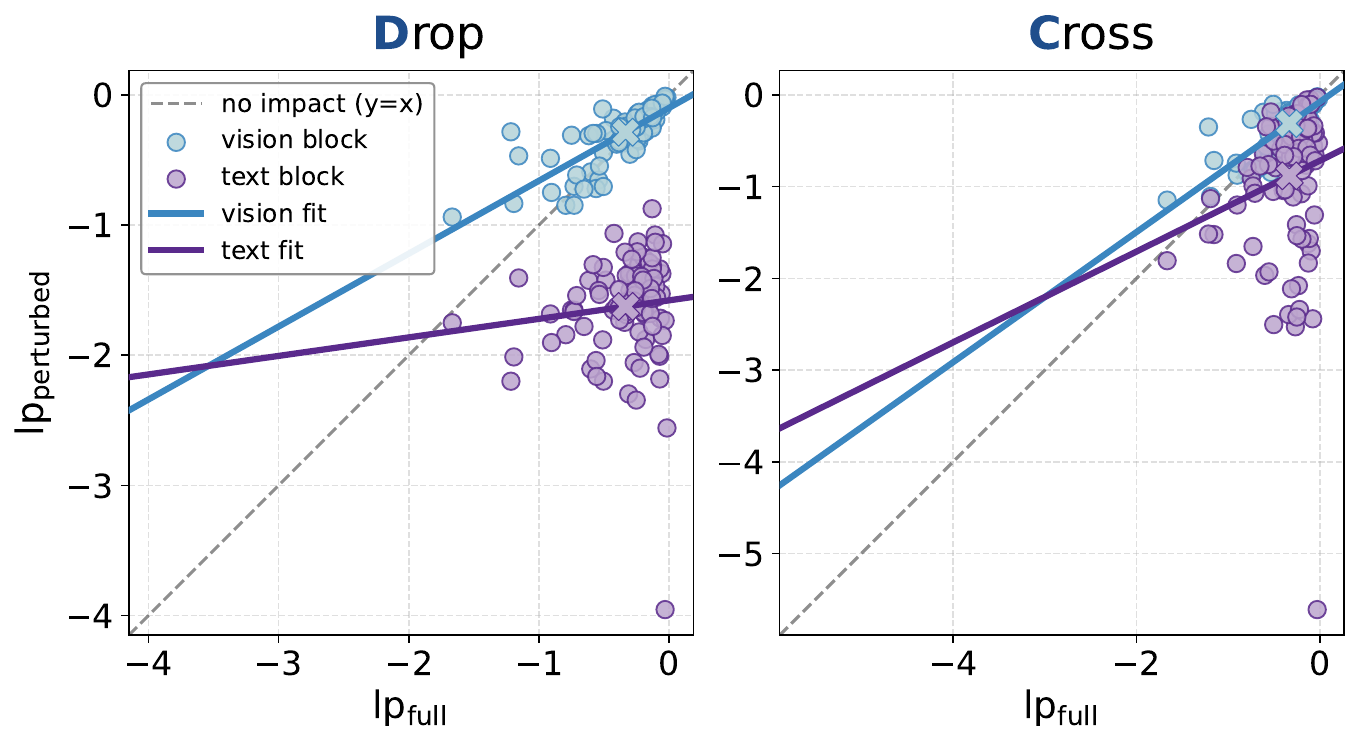}
    \caption{
    \textbf{Per-sample vision vs.\ text perturbation impact for Monet-7B on MMVP $+$ V\textsuperscript{*}Bench.}
    Monet emits a single visual latent block per trajectory, so vision-side \texttt{Shuffle} is undefined and omitted. \texttt{Drop} and \texttt{Cross} are evaluated on all $100$ samples. Since Monet has no between-block text segments, text-side \texttt{Drop} uses the naive full-text variant. Both vision and text channels carry signal in this single-hop setting, but text perturbations cause larger log-probability drops.
    }
    \label{fig:monet_scatter}
\end{figure}

This section reports per-sample perturbation plots for additional latent-reasoning baselines. As shown in Figures~\ref{fig:skila_scatter}, \ref{fig:covt_scatter}, and \ref{fig:monet_scatter}, the same text-over-vision asymmetry observed for Mirage-7B also appears in SkiLa-7B, CoVT-7B, and Monet-7B. These results complement the aggregate perturbation results in the main text and show that Mental Simulation Collapse is not specific to a single model architecture or benchmark.

\section{Training and Inference Cost}
\label{app:cost}
\textsc{AnchorSim} adds no inference overhead: the visual-blind branch and the EMA teacher are used only during training and are discarded after Stage 2, so \textsc{AnchorSim-7B} has the same architecture, parameter count, and latent format as the Stage 1 model. During training, the two auxiliary passes do not require gradients, so neither requires a backward pass or a separate training run.

\section{Why Do Models with MSC Still Improve?}
\label{app:msc_improve}
We found baselines with MSC can still improve over the base model, plausibly through channels other than the content of the visual latents. Latent blocks add serial computation, like pause tokens~\cite{goyal2023think}; they structure the reasoning trajectory into multi-step segments~\cite{li2025llms, wang2023towards}; and Stage 1 alignment with helper-image embeddings may change how the model attends to the input image. Our diagnostics do not isolate these effects, but all three are consistent with MSC: these models may work in spite of their mental images rather than because of them. \textsc{AnchorSim} preserves these structural benefits while also making the latent content consequential.

\section{Societal Impact}
\label{app:societal_impact}

This work aims to make multimodal reasoning systems more reliable by ensuring that generated visual latents are not merely decorative, but are actually used in downstream prediction. By mitigating \emph{Mental Simulation Collapse}, AnchorSim may improve applications that require faithful reasoning over visual inputs, including assistive technologies, educational tools, scientific diagram understanding, spatial planning, and analysis of complex visual content. More causally grounded visual reasoning may also improve interpretability, since perturbation-based diagnostics can help determine whether a model relies on visual evidence rather than unsupported linguistic priors.

At the same time, stronger multimodal reasoning systems introduce risks. Models that better understand images and visual structure may be misused for privacy-invasive analysis, surveillance, or automated decision-making in sensitive domains. They may also inherit or amplify biases present in training data, especially when visual cues correlate with demographic or social attributes. AnchorSim addresses one failure mode of latent multimodal reasoning, but it does not guarantee fairness, privacy, or safety. Responsible deployment requires careful dataset curation, bias evaluation, privacy safeguards, human oversight, and domain-specific validation before use in high-stakes settings.

\clearpage
\section{Additional Qualitative Examples}
\label{app:qualitative_examples}
\begin{figure}[t]
    \centering
    \includegraphics[width=\linewidth]{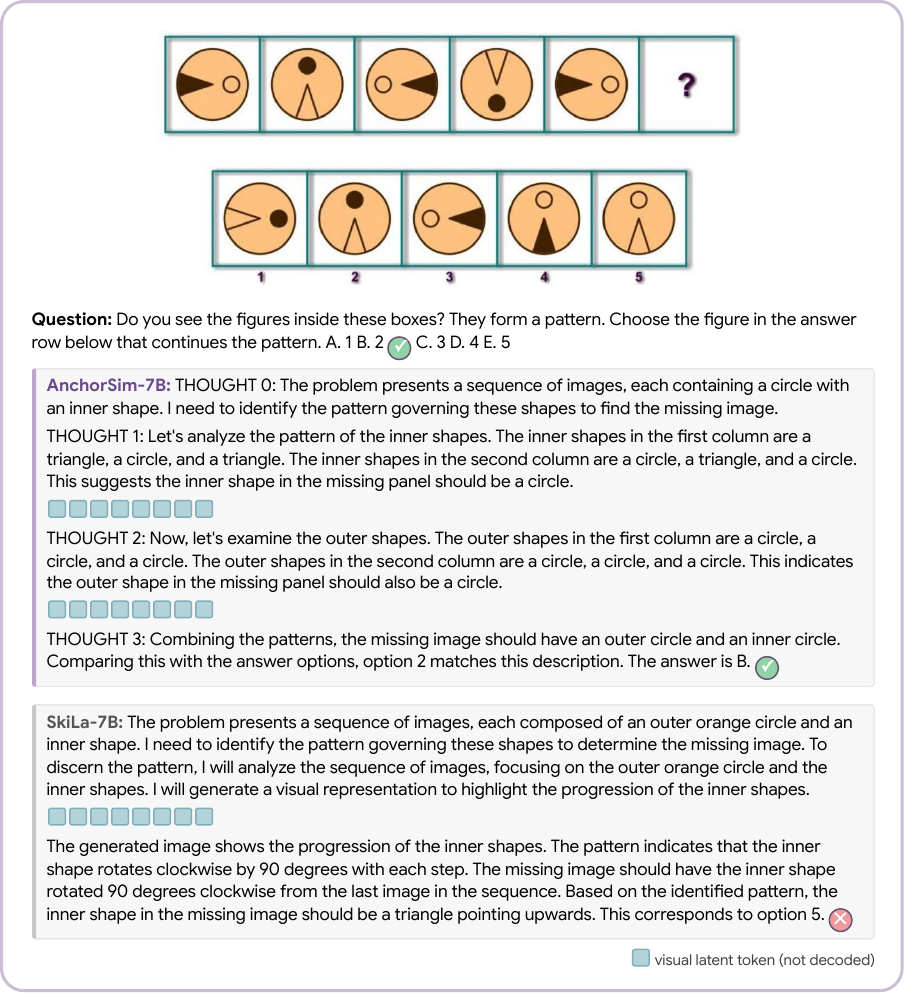}
    \caption{
    \textbf{Qualitative example on EMMA.}
    The model solves a diagram-grounded STEM problem by combining textual reasoning with visual grounding. AnchorSim encourages the reasoning chain to revisit visual latents when deriving the final answer, rather than relying only on language priors.
    }
    \label{fig:emma}
\end{figure}

\begin{figure}[t]
    \centering
    \includegraphics[width=\linewidth]{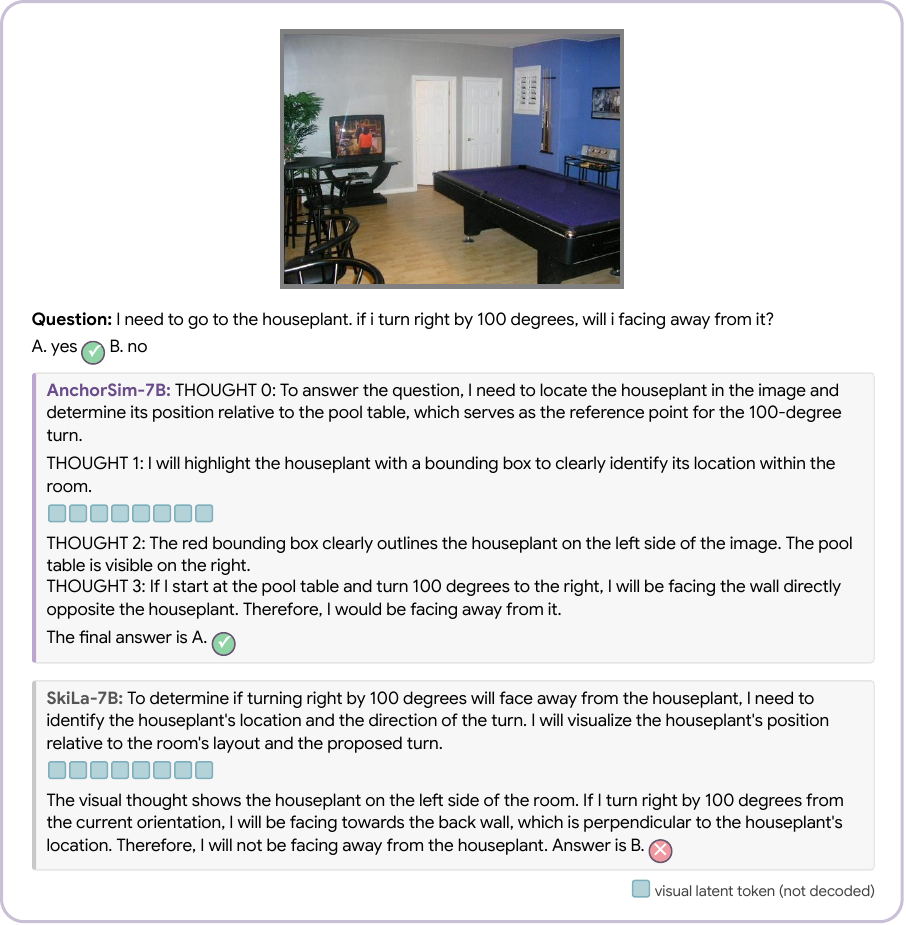}
    \caption{
    \textbf{Qualitative example on SAT-R.}
    The task requires mentally simulating a spatial or viewpoint transformation from a real image. The example illustrates how latent visual states can support intermediate spatial reasoning before the model commits to the final answer.
    }
    \label{fig:sat_1}
\end{figure}

\begin{figure}[t]
    \centering
    \includegraphics[width=\linewidth]{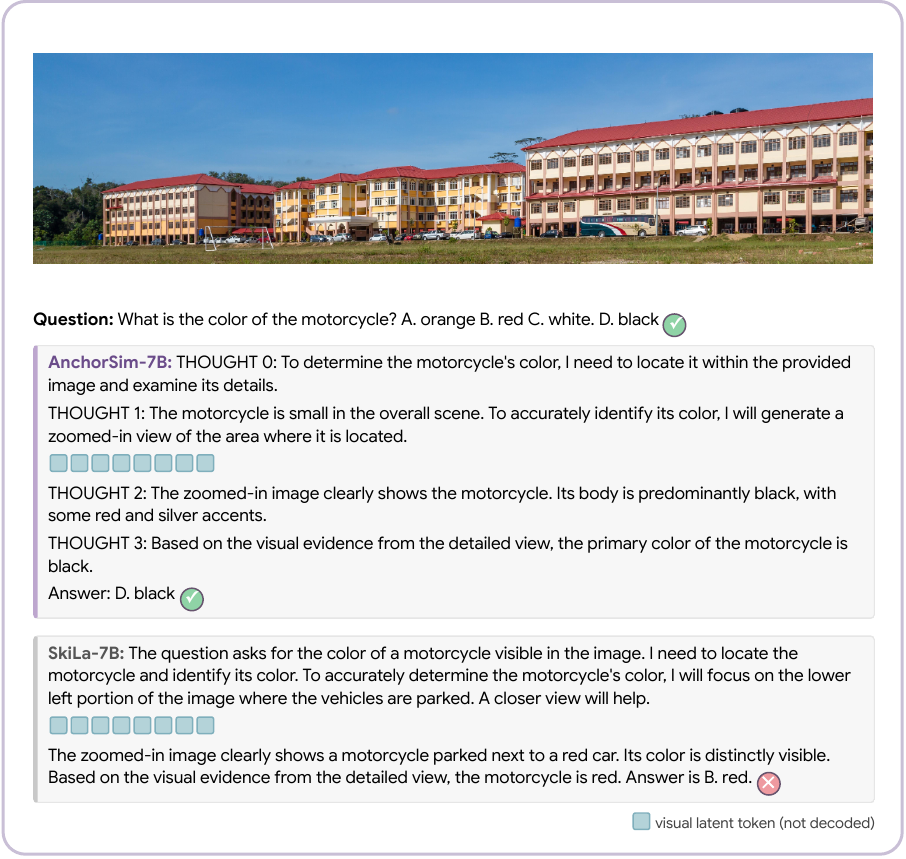}
    \caption{
    \textbf{Qualitative example on V\textsuperscript{*}.}
    The question depends on fine-grained visual search over a high-resolution image. AnchorSim helps the model preserve visually relevant details in latent form, allowing the final response to be grounded in localized evidence.
    }
    \label{fig:vstar}
\end{figure}

\begin{figure}[t]
    \centering
    \includegraphics[width=\linewidth]{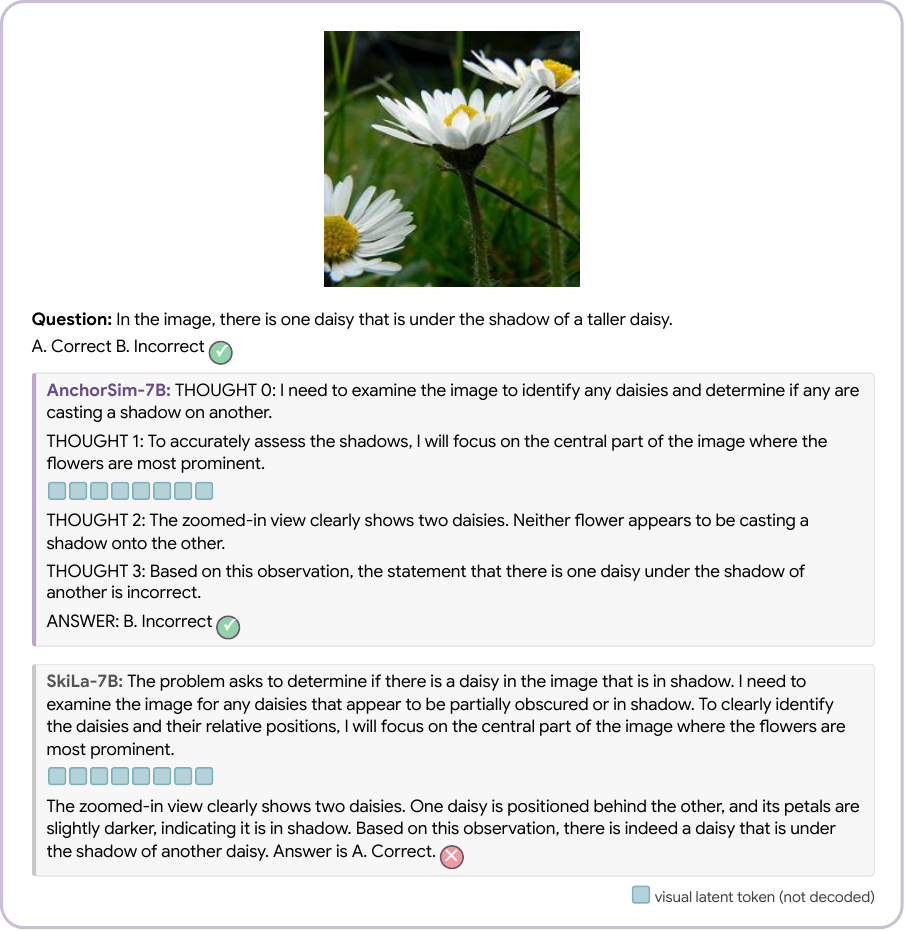}
    \caption{
    \textbf{Qualitative example on MMVP.}
    MMVP contains image pairs that expose failures in low-level visual discrimination. This example shows how anchoring the reasoning process to visual latents can help distinguish visually similar alternatives.
    }
    \label{fig:mmvp}
\end{figure}

\begin{figure}[t]
    \centering
    \includegraphics[width=\linewidth]{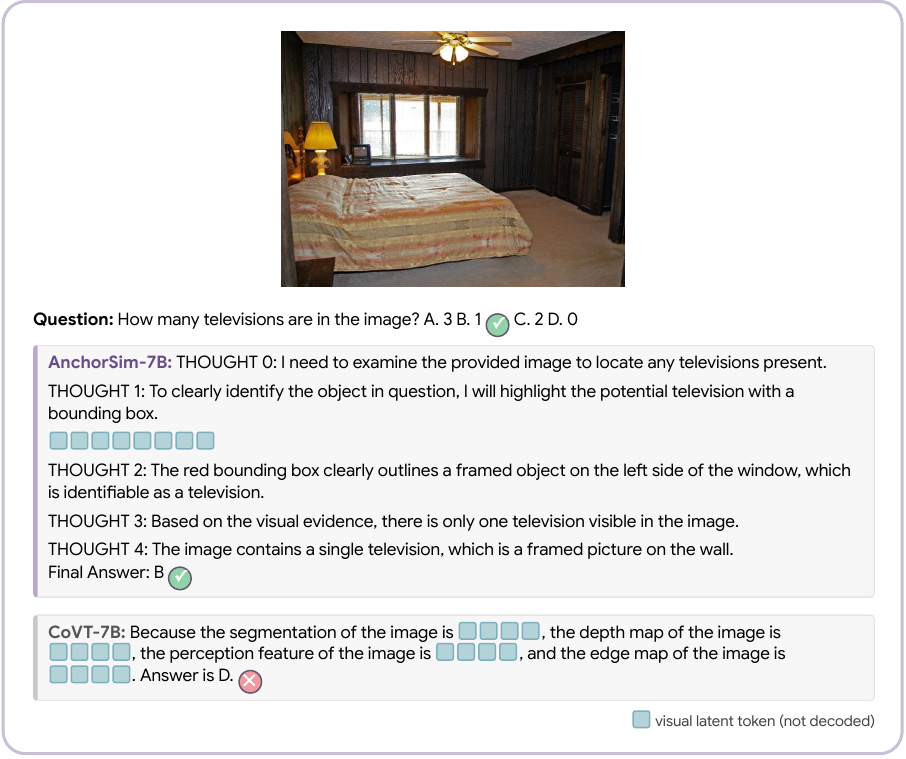}
    \caption{
    \textbf{Qualitative example on CV-Bench-2D.}
    The model must reason about planar spatial relations and object layout. The generated latent states provide compact visual anchors that support relation tracking throughout the reasoning trajectory.
    }
    \label{fig:cvbench}
\end{figure}

\begin{figure}[t]
    \centering
    \includegraphics[width=\linewidth]{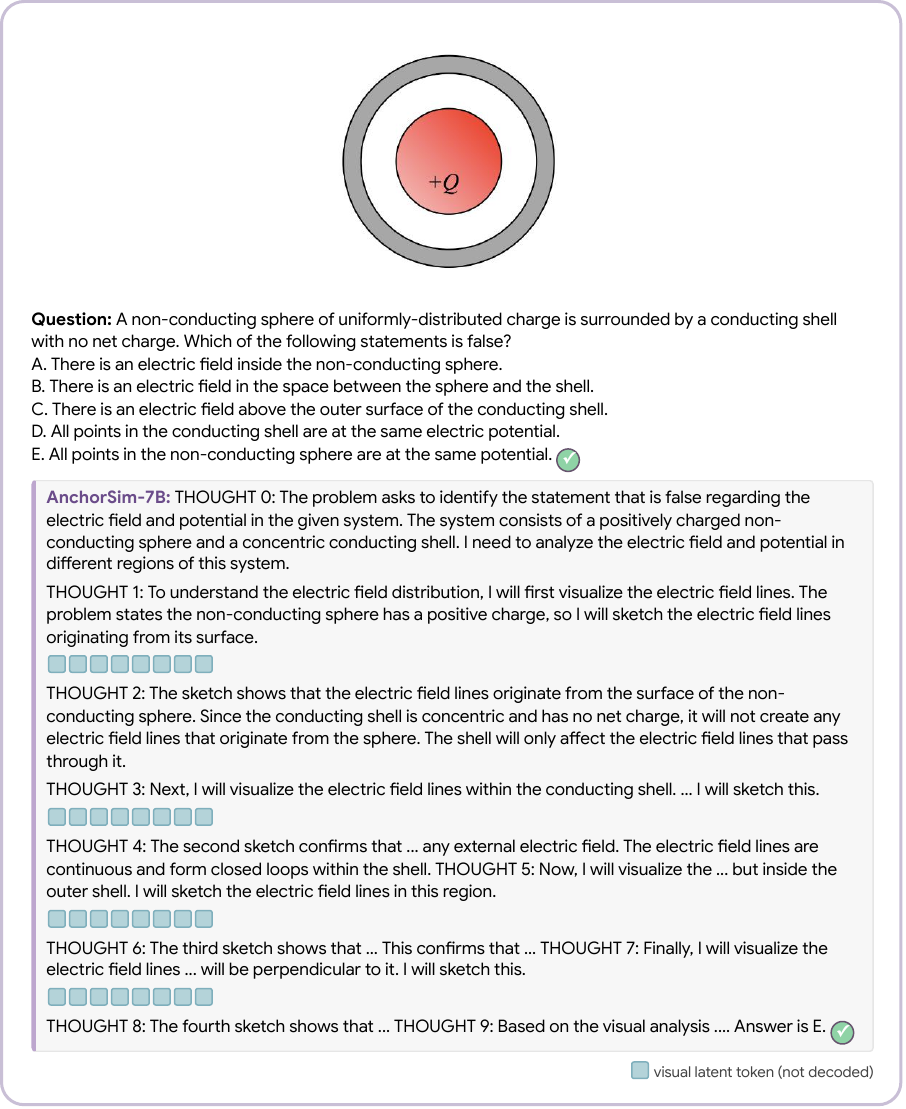}
    \caption{
    \textbf{Qualitative example of multi-hop visual reasoning.}
    The model decomposes the problem into multiple intermediate reasoning steps, each paired with latent visual states. This example illustrates the intended behavior of interleaved visual reasoning: later text should depend on earlier visual simulations rather than bypassing them.
    }
    \label{fig:multi-hop}
\end{figure}

We provide additional qualitative examples to illustrate how AnchorSim behaves across different types of visual reasoning tasks. Figure~\ref{fig:emma} shows a diagram-grounded STEM example from EMMA, where the model must combine symbolic reasoning with visual evidence. Figure~\ref{fig:sat_1} demonstrates spatial mental simulation on SAT-Real, while Figure~\ref{fig:vstar} highlights fine-grained visual search on V\textsuperscript{*}. Figure~\ref{fig:mmvp} presents a low-level visual discrimination case from MMVP, and Figure~\ref{fig:cvbench} shows 2D spatial-relation reasoning on CV-Bench. Finally, Figure~\ref{fig:multi-hop} illustrates a multi-hop reasoning trajectory in which intermediate latent visual states support later textual reasoning before the final answer.

\end{document}